\documentclass[11pt]{article}

\usepackage[preprint]{acl}

 \usepackage{microtype}

\usepackage[english]{babel}
\usepackage{fontspec}
\usepackage{graphicx}
\usepackage{booktabs}
\usepackage{multirow}
\usepackage{amsmath}
\usepackage{amssymb}
\usepackage{algorithm}
\usepackage{algorithmic}

\title{SkillEval: Decomposing Agent Skill Quality into Interpretable Signals}

\author{
Jiahui Han\textsuperscript{1,2,*},
Qinuo Li\textsuperscript{2,5,*},
Ziheng Peng\textsuperscript{2,4,*},
Haotian Wu\textsuperscript{3},
Haoze Liu\textsuperscript{2},\\
\bfseries Danfeng Shan\textsuperscript{1},
Guanchu Wang\textsuperscript{2},
Huiqi Deng\textsuperscript{1,2,\textdagger},
Ninghao Liu\textsuperscript{3,\textdagger}\\[0.5em]
{\normalfont\normalsize\textsuperscript{1}Xi'an Jiaotong University}\\
{\normalfont\normalsize\textsuperscript{2}Shanghai AI Laboratory}\\
{\normalfont\normalsize\textsuperscript{3}The Hong Kong Polytechnic University}\\
{\normalfont\normalsize\textsuperscript{4}Renmin University of China}\\
{\normalfont\normalsize\textsuperscript{5}Harbin Institute of Technology}
}

\begin{document}

\maketitle
\begin{abstract}
Agent skills provide reusable knowledge that helps agents solve specialized tasks. As their use expands, evaluating skill quality becomes increasingly important. Existing evaluations often measure skill quality based on improvements in agent performance on specific tasks. However, a reusable skill may apply across multiple scenarios. Evaluation on a single downstream task mainly reflects the compatibility between a skill and the evaluated task, provides only a partial view of skill quality, and does not identify which aspects need improvement.
We find that general properties of the \texttt{SKILL.md} document play an important role in skill quality. To evaluate these properties, we propose \textbf{SkillEval}, an interpretable framework for document-level skill evaluation. SkillEval produces interpretable scores for each property using a fixed, inspectable scoring direction. It further measures and reduces the influence of unrelated document features, so each score more specifically captures its intended semantic property.
We use SkillEval to evaluate skills in controlled quality tests, showing that it reliably distinguishes skills of different quality. Its scores closely reflect downstream task performance, providing an early indication of skill effectiveness. We further explore SkillEval for diagnosing document weaknesses and guiding targeted revisions that improve the targeted properties and increase downstream task pass rates.
\end{abstract}

\section{Introduction}


Agent skills increasingly serve as reusable procedural documents that provide agents with task-specific knowledge, tool-use guidance, and procedural workflows at inference time~\citep{jiang2026sokagenticskills,
ling2026agentskillsdatadrivenanalysis}. A skill packages the recurring procedure for a class of tasks into a form that can be stored, retrieved, and reused without retraining the underlying model~\citep{agentskills_specification}. This makes skills a practical mechanism for accumulating experience in modern agent systems
~\citep{liang2026skillnet,lei2026skillevolbench}. As agents are deployed in increasingly specialized tasks, skills are becoming a common way to extend agents beyond their base capabilities~\citep{jiang2026sokagenticskills,
ling2026agentskillsdatadrivenanalysis,
li2026skillsbenchbenchmarkingagentskills}. \looseness=-1

As skills are increasingly used, the evaluation of skill quality becomes increasingly important. The quality of a skill directly affects how well an agent can understand the task and complete the required workflow. Recent work has studied the evaluation of skills through the outcomes of downstream tasks, such as task pass rates, or through analyses of agent execution trajectories ~\citep{li2026skillsbenchbenchmarkingagentskills,
han2026swe,
zhong2026skilllearnbench}.
However, existing evaluations still test skills in specific downstream task environments. 
This makes the observed performance highly dependent on the compatibility between a skill and the evaluated task~\citep{liu2026agenticskillsworkwild}. In addition, the pass rate alone cannot identify which aspect of a skill causes poor performance or how the skill should be improved. For a reusable skill, quality is not determined only by its performance on one task, but also by its ability to support different task scenarios consistently. Therefore, single-task downstream performance provides a useful but partial view of skill quality.

We find that the general properties of a \texttt{SKILL.md} document also play an important role in determining quality of skills. To evaluate these document-level properties, we propose SkillEval, an interpretable framework that evaluates the skill document. SkillEval decomposes skill quality into four complementary dimensions: applicability, content quality, execution guidance, and robustness, as illustrated in Table~\ref{tab:skill_metrics}. These dimensions examine when a skill should be used, whether its content is clear and specific, how it guides execution, and how it handles failures and edge cases. In addition, for each learned metric, SkillEval derives an explicit direction from controlled positive--negative skill pairs and scores a document by projecting its representation onto that direction. This fixed scoring procedure makes individual quality signals inspectable and reproducible. It also enables potential biases introduced by document length or formatting to be quantified and corrected by orthogonalizing the metric direction against the corresponding bias direction in hidden space.

\begin{table*}[t]
\centering
\small
\setlength{\tabcolsep}{4pt}
\renewcommand{\arraystretch}{1.05}
\renewcommand{\multirowsetup}{\centering}
\hyphenpenalty=10000
\begin{tabular}{@{}p{0.17\textwidth}p{0.31\textwidth}p{0.48\textwidth}@{}}
\toprule
\multicolumn{1}{c}{\textbf{Dimension}}
& \textbf{Metric Name}
& \textbf{Definition} \\
\midrule

\multirow{2}{=}{\textbf{Applicability}}
& \textbf{A1}\; Format Validity
& Whether the skill is a Markdown file with valid YAML frontmatter and the required \texttt{name} and \texttt{description} fields. \\

& \textbf{A2}\; Trigger Clarity
& Whether the description states what the skill does, when it should be used, and its main applicability conditions. \\

\midrule

\multirow{2}{=}{\textbf{Content Quality}}
& \textbf{B1}\; Content Consistency
& Whether the capabilities stated in the description are supported by the instructions and resources in the body. \\

& \textbf{B2}\; Technical Specificity
& Whether the skill provides concrete, executable details, including tools, commands, parameters, fields, units, constraints, and file paths. \\

\midrule

\multirow{2}{=}{\textbf{Execution Guidance}}
& \textbf{C1}\; Workflow Integrity
& Whether the skill specifies the required input, processing, and output stages. \\

& \textbf{C2}\; I/O Explicitness
& Whether the expected inputs and outputs, formats, fields, schemas, units, and constraints are explicitly defined. \\

\midrule

\multicolumn{1}{c}{\textbf{Robustness}}
& \textbf{D1}\; Failure Awareness
& Whether the skill covers common failures, boundary conditions, diagnostic signals, and recovery strategies. \\

\bottomrule
\end{tabular}
\caption{Intrinsic skill-quality properties used in our evaluation framework.}
\label{tab:skill_metrics}
\end{table*}

We use SkillEval to evaluate skills and analyze their downstream utility on \textsc{SkillsBench}~\citep{li2026skillsbenchbenchmarkingagentskills}, leading to three key findings:
\textit{
(1) The learned metric directions distinguish positive skills from negative skills across A2, B1, B2, C1, C2, and D1 on the validation set of 518 skills, including 259 positive and 259 negative examples. 
(2) The predictions of pass-rate uplift, defined as the difference between the pass rate with and without a skill, are strongly correlated with the observed uplift on \textsc{SkillsBench}~\citep{li2026skillsbenchbenchmarkingagentskills}, with Pearson $r$ ranging from \textbf{0.779} to \textbf{0.787}.
(3) SkillEval identifies weaknesses in skill documents and guides LLMs to address them. These targeted revisions improve the relevant metric scores and increase the mean downstream pass rate by \textbf{29.5} percentage points.}
These findings demonstrate that SkillEval provides interpretable signals that reflect both document-level quality and downstream skill utility.

Our main contributions are as follows:
\begin{itemize}
    \item We propose \textbf{SkillEval}, an interpretable framework that decomposes skill quality into four complementary
        dimensions. 

    \item We show that downstream pass-rate uplift is strongly correlated with \textbf{SkillEval} scores, providing an
        early indication of whether a skill can help an agent complete a task.
    \item We explore \textbf{SkillEval} as a tool for diagnosing and improving skills. These metric-level scores
      identify specific weaknesses and guide targeted revisions of skill documents. Our experiments show that the
      revised skills improve the targeted properties and improve pass rates on downstream tasks.
  \end{itemize}

\section{Related Work}
Agent skills package instructions, workflows, and resources for reuse
~\citep{liang2026skilltextskillstructure}. Recent surveys organize this
emerging field and discuss systematic evaluation
~\citep{jiang2026sokagenticskills,
ding2026agentskillevaluationevolution}. We group prior work by its source
of evaluation evidence. Task-based methods infer skill quality from
downstream execution, whereas intrinsic methods assess the skill
document without executing a downstream task. They differ in cost,
environment dependence, and the evidence they provide.  \looseness=-1

\subsection{Task-Based Skill Evaluation}

Task-based evaluation measures whether a skill improves an agent on
downstream tasks. 
\textsc{SkillsBench}~\citep{li2026skillsbenchbenchmarkingagentskills}
compares matched runs with and without a
curated skill. It uses containerized environments and deterministic
verifiers across several domains
~\citep{li2026skillsbenchbenchmarkingagentskills}. SWE-Skills-Bench uses a
similar design for software engineering. It pairs public skills with
pinned repositories and executable acceptance tests
~\citep{han2026swe}. These benchmarks provide direct evidence
of practical utility. \looseness=-1

Other work broadens the scope of task-based evaluation
~\citep{zhou2026skillgenbenchbenchmarkingskillgeneration,ying2026openskillevalautomaticallyauditingopen}.
SkillLearnBench
separates skill-text quality, trajectory alignment, and final task
outcomes ~\citep{zhong2026skilllearnbench}. Some studies generate
tasks and scoring rubrics from each skill
~\citep{shaposhnikov2026framework}. AgentSkillOS evaluates skill retrieval
and composition over large libraries, with pairwise LLM evaluation of
task outputs ~\citep{li2026agentskillos}. These systems cover more stages
of skill use than a single success score~\citep{zhu2026skillcoachselfevolvingrubricsevaluating}.

\begin{figure*}[t]
    \centering
    \includegraphics[width=\textwidth]{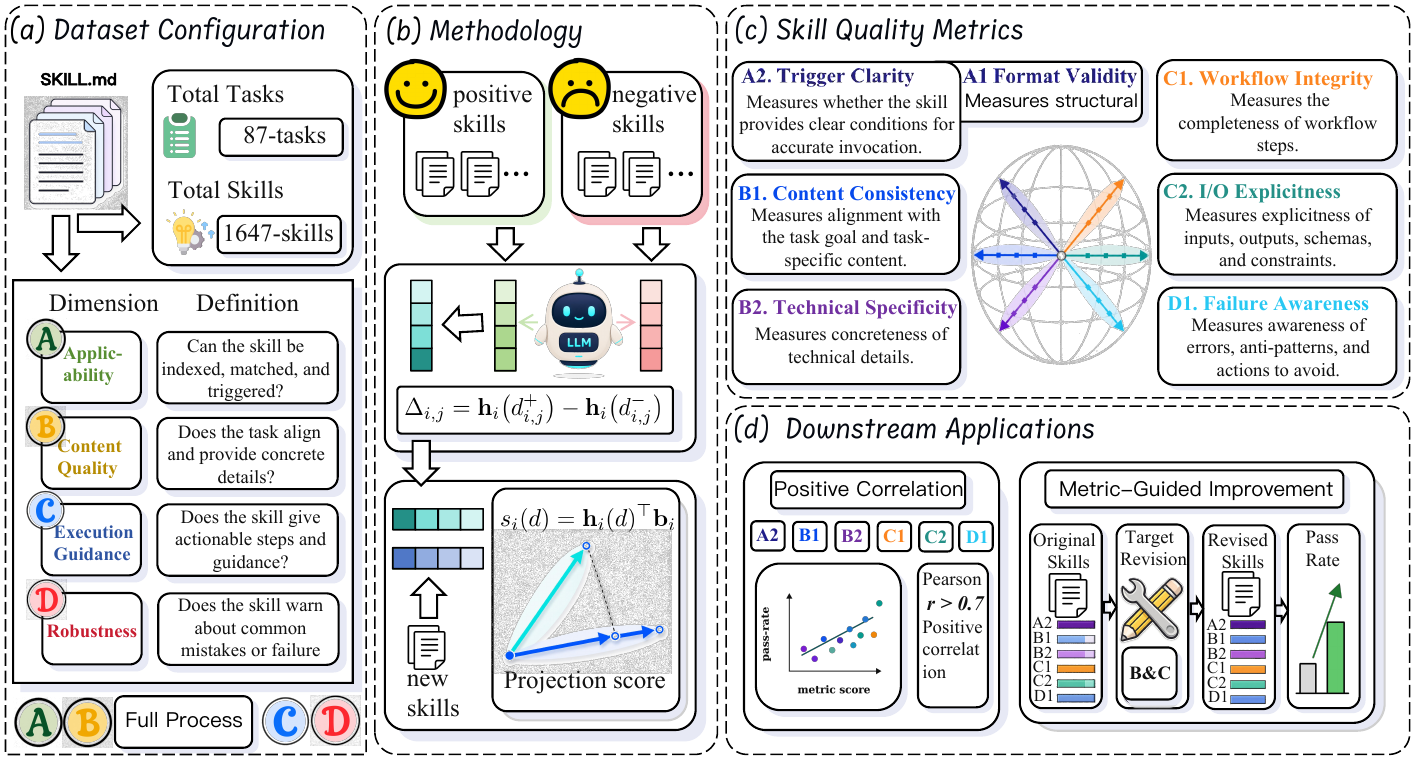}
    \caption{Overview of SkillEval. The figure shows (a) the dataset configuration, where 1,647 \texttt{SKILL.md} documents from 87 tasks are organized into four
  quality dimensions; (b) the methodology, where strong and weak skills are contrasted in a frozen LLM representation space to learn metric-specific
  scoring directions. Each learned direction assigns a score to the corresponding quality dimension of a skill document; (c) the skill quality metrics used to evaluate reusable skill documents; and (d) the downstream applications, where the learned
  scores are used for pass-rate correlation analysis and metric-guided skill revision.}
    \label{fig:skillsMain}
\end{figure*}
Despite these differences, task-based methods require agent execution.
They need task environments and outcome verification. Many also require
tool access.
Their results can also depend on the base model, agent harness, and
resource budget ~\citep{li2026skillsbenchbenchmarkingagentskills,
xu2026agent}. Task-level rewards may mix skill quality
with agent capability ~\citep{ding2026agentskillevaluationevolution}.
Therefore, this approach is costly for frequent screening. A failed run
may also reflect the skill, model, tool, or agent setup.\looseness=-1

\subsection{Intrinsic Skill Quality Evaluation}
Agent skills exhibit observable document-level properties, and these
properties can affect how skills are discovered, selected, and governed
in practice
~\citep{ling2026agentskillsdatadrivenanalysis,saha2026under}.
Existing work has begun to assess such intrinsic properties directly.
\citet{hong2026anatomy} treat \texttt{SKILL.md} as a software artifact,
define skill smells such as weak guidance, missing safeguards, and
context bloat, and develop methods to detect them.
SkillLearnBench evaluates skill-text quality alongside trajectory
alignment and task outcomes, including text-level coverage,
executability, and safety
~\citep{zhong2026skilllearnbench}.

LLM judges are useful when explicit rules are hard to write. MT-Bench
and G-Eval show that strong judges can align well with human judgments
in dialogue and text evaluation
~\citep{zheng2023judging,liu-etal-2023-g}. However, their reliability
depends on the evaluation setting. Candidate order can change pairwise
judgments ~\citep{shi2025judging}, and non-target information can affect
decisions ~\citep{chen2024humans}. Reliability also varies across models,
tasks, and evaluated properties ~\citep{bavaresco2025llms}. Code judges
can change scores after surface edits that preserve program behavior
~\citep{moon2026don}. Therefore, repeated generative judgments may change
with the judge or the presentation of the input.\looseness=-1

\subsection{Internal Representations for Evaluation}
Prior work suggests that high-level concepts can be represented as
approximately linear directions in a model's internal representation
space
~\citep{zou2023representation,kim2018interpretability,park2023linear}.
Such directions can be extracted from contrastive examples and used to
analyze or influence model behavior
~\citep{rimsky-etal-2024-steering,NEURIPS2023_81b83900}.
Recent studies further show that internal representations can serve as
evaluation signals through learned probes or projections onto
interpretable semantic directions
~\citep{li2026rethinking,ma-etal-2026-stable}.
To our knowledge, however, prior work has not applied this paradigm to
the fine-grained quality evaluation of agent skill documents.\looseness=-1


Motivated by this gap, we develop an interpretable framework
for evaluating the intrinsic quality of agent skills. SkillEval learns one
representation direction for each metric. At evaluation time, the
directions remain fixed, and the framework produces dimension-level
scores without querying a generative judge.\looseness=-1

\section{Proposed Methodology}

We propose an interpretable framework for evaluating the general properties of skills across four complementary dimensions: applicability, content quality, execution guidance, and robustness. As shown in Figure~\ref{fig:skillsMain}, these dimensions are represented by seven metrics, comprising one structural-format metric and six semantic metrics. Structural Format Validity (A1) is evaluated using deterministic rules. For each semantic metric, we construct positive--negative skill pairs that differ primarily in the target property. We encode the paired skills with a frozen language model and aggregate their representation differences to derive a metric-specific quality direction. At evaluation time, we project a new skill onto each quality direction and standardize the resulting scores using
training-split statistics. Together with the structural check, these scores form an interpretable quality profile for each skill. \looseness=-1

\subsection{Metric Selection and Definitions}

Agent skills support diverse tasks, domains, and tool environments
~\citep{jiang2026sokagenticskills,
li2026skillsbenchbenchmarkingagentskills}.
Therefore, task-specific evaluation criteria may not apply uniformly
across skills
~\citep{ding2026agentskillevaluationevolution}.
We focus instead on intrinsic properties that can be evaluated
directly from the skill document. We select metrics that are broadly
applicable, observable from \texttt{SKILL.md}, and related to common
sources of skill failure. Table~\ref{tab:skill_metrics} summarizes the
selected metrics.


Structural Format Validity is checked with deterministic
rules. The remaining metrics require semantic judgments and are learned
from controlled skill pairs as described below. \looseness=-1

\subsection{Positive--Negative Skill Pair Construction}

A reliable metric direction requires positive and negative examples that
differ mainly in the target quality dimension. If the two skills also
differ in topic, task content, writing style, or length, the learned
direction may capture these unrelated changes. Therefore, we use multiple
controlled skill pairs to reduce noise from individual examples
.\looseness=-1

For each learned semantic metric $F_i$, we first select a positive skill
$d_{i,j}^{+}$ that meets the metric criteria. We then select its negative
version with a metric-specific selection function $T_i$:\looseness=-1
\[
d_{i,j}^{-}=T_i(d_{i,j}^{+}).
\]
The selected counterpart weakens the target property while keeping the task goal, topic,
tools, and other content as unchanged as possible. This design
limits unrelated differences and helps the learned direction focus on the
target metric.

\subsection{Hidden Representation and Direction Extraction}

We use a frozen Qwen3.5-9B model and extract the hidden-state tensor at
index 16. For metric $F_i$ and skill $d$, let
\[
\mathbf{H}^{(16)}_i(d)\in\mathbb{R}^{T_d\times d_h}
\]
denote the token-level hidden states. We pool the region most relevant to
each metric:
\[
\mathbf{h}_i(d)
=
\mathrm{Pool}_i\!\left(\mathbf{H}^{(16)}_i(d)\right)
\in\mathbb{R}^{d_h}.
\]
A2 uses the frontmatter description. B1 averages the description and body
representations. B2, C1, and C2 use mean pooling over the body. D1 uses
the final token of the full model input. The model, hidden-state index,
prompt, and pooling rule are fixed before validation. The maximum input
length is 20,000 tokens, and no document in our dataset is truncated.\looseness=-1

For the $j$-th training pair, we compute
\[
\Delta_{i,j}
=
\mathbf{h}_i\!\left(d_{i,j}^{+}\right)
-
\mathbf{h}_i\!\left(d_{i,j}^{-}\right).
\]

A confounding factor may introduce an unintended direction in the hidden
space. Following prior work on removing linearly represented attributes
through projection
~\citep{ravfogel2020null,haghighatkhah2022better,belrose2023leace}, we use a precomputed
unit confounding direction $\mathbf{u}_{\mathrm{conf}}$, fixed before
metric training and shared across semantic metrics. We remove the
component aligned with this direction from each training difference:\looseness=-1

\[
\Delta_{i,j}^{\perp}
=
\Delta_{i,j}
-
\left(
\Delta_{i,j}^{\top}\mathbf{u}_{\mathrm{conf}}
\right)
\mathbf{u}_{\mathrm{conf}}.
\]
The metric direction is the normalized mean of the orthogonalized
training differences:
\[
\mathbf{b}_i
=
\frac{
\frac{1}{N_i}\sum_{j=1}^{N_i}\Delta_{i,j}^{\perp}
}{
\left\|
\frac{1}{N_i}\sum_{j=1}^{N_i}\Delta_{i,j}^{\perp}
\right\|_2
}.
\]
\begin{algorithm}[t]
\caption{SkillEval Direction Learning and Scoring}
\label{alg:skilleval}
\begin{algorithmic}[1]
\REQUIRE Paired training sets
$\mathcal{P}_i=\{(d_{i,j}^{+},d_{i,j}^{-})\}_{j=1}^{N_i}$,
$i=1,\ldots,K$,
Pooled representation functions
$\{\mathbf{h}_i(\cdot)\}_{i=1}^{K}$,
unit confounding direction $\mathbf{u}_{\mathrm{conf}}$,
unseen skill $d$
\ENSURE Metric directions $\{\mathbf{b}_i\}_{i=1}^{K}$
and standardized scores $\{z_i(d)\}_{i=1}^{K}$
\FOR{$i=1,\ldots,K$}
    \STATE $\mathcal{D}_i^{\mathrm{train}}
    \gets
    \bigcup_{j=1}^{N_i}
    \{d_{i,j}^{+},d_{i,j}^{-}\}$

    \FOR{$j=1,\ldots,N_i$}
        \STATE $\Delta_{i,j}\gets
        \mathbf{h}_i(d_{i,j}^{+})
        -\mathbf{h}_i(d_{i,j}^{-})$
        \STATE $\Delta_{i,j}^{\perp}\gets
        \Delta_{i,j}
        -
        (\Delta_{i,j}^{\top}\mathbf{u}_{\mathrm{conf}})
        \mathbf{u}_{\mathrm{conf}}$
    \ENDFOR

    \STATE $\bar{\Delta}_i\gets
    \frac{1}{N_i}
    \sum_{j=1}^{N_i}\Delta_{i,j}^{\perp}$

    \STATE $\mathbf{b}_i\gets
    \bar{\Delta}_i/
    \lVert\bar{\Delta}_i\rVert_2$

    \STATE $\mathcal{S}_i^{\mathrm{train}}\gets
    \left\{
    \mathbf{h}_i(\tilde d)^{\top}\mathbf{b}_i
    \mid
    \tilde d\in\mathcal{D}_i^{\mathrm{train}}
    \right\}$

    \STATE $(\mu_i^{\mathrm{train}},
    \sigma_i^{\mathrm{train}})
    \gets
    \operatorname{MeanStd}
    (\mathcal{S}_i^{\mathrm{train}})$

    \STATE $s_i(d)\gets
    \mathbf{h}_i(d)^{\top}\mathbf{b}_i$

    \STATE $z_i(d)\gets
    \dfrac{s_i(d)-\mu_i^{\mathrm{train}}}
    {\sigma_i^{\mathrm{train}}}$
\ENDFOR

\STATE \textbf{return}
$\{\mathbf{b}_i\}_{i=1}^{K},
\{z_i(d)\}_{i=1}^{K}$
\end{algorithmic}
\end{algorithm}
\subsection{Metric Scoring and Normalization}
For a new skill $d$, we compute the metric score by projection:\looseness=-1
\[
s_i(d)=\mathbf{h}_i(d)^{\top}\mathbf{b}_i.
\]
We convert the score to a $z$-score using the mean and standard
deviation over all positive and negative training skills:
\[
z_i(d)
=
\frac{s_i(d)-\mu_i^{\mathrm{train}}}
{\sigma_i^{\mathrm{train}}}.
\]
Higher values indicate stronger alignment with the positive side of
metric $F_i$.

\section{Experiments}

\subsection{Experimental Setup}

\noindent\textbf{Dataset Setting.}
Our dataset contains 1,647 skills from open-source datasets, public online sources, manual construction, and LLM generation. For each semantic metric, we construct controlled positive--negative pairs and split them into training and validation sets, with no overlap in original documents, tasks, sources, or generation templates. All metric directions and normalization statistics are learned exclusively from the training split, while the validation split is used only for final evaluation. \looseness=-1

\noindent\textbf{Model Setting.}
In our main experiments, we use a frozen Qwen3.5-9B model and extract representations at hidden-state index 16. A2
applies mean pooling to the description, B1 combines the mean-pooled representations of the description and body,
B2, C1, and C2 apply mean pooling to the body, and D1 uses the final-token representation. The maximum input length
is 20,000 tokens. For model ablations, we further use Qwen3.5-27B at index 32, Llama3.1-8B-Instruct at index 16, and
Llama3.2-3B-Instruct at index 14.

\noindent\textbf{Downstream Evaluation.}
We use \textsc{SkillsBench}~\citep{li2026skillsbenchbenchmarkingagentskills} for downstream evaluation and run each
task with Codex using GPT-5.5 at the xhigh reasoning effort. For each task, the with-skill and no-skill conditions
use the same task environment and executor configuration. Each condition is evaluated over three runs, and the pass
rate is calculated as the proportion of successful runs. We define pass-rate uplift as the difference between the
pass rates under the with-skill and no-skill conditions. \looseness=-1

\begin{figure}[t]
  \centering
  \includegraphics[width=1.0\columnwidth]{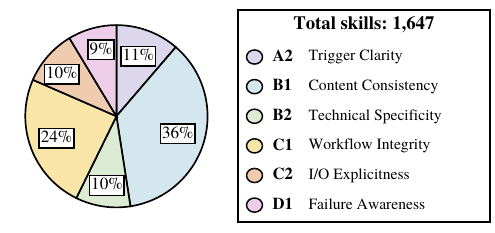}
 \caption{Distribution of the 1,647 skills used to construct and validate the six metric directions: A2, B1, B2, C1, C2, and D1.}
  \label{fig:metric_skill_distribution}
\end{figure}
\subsection{Main Result}
\begin{figure}[t]
  \centering
  \includegraphics[width=1.0\columnwidth]{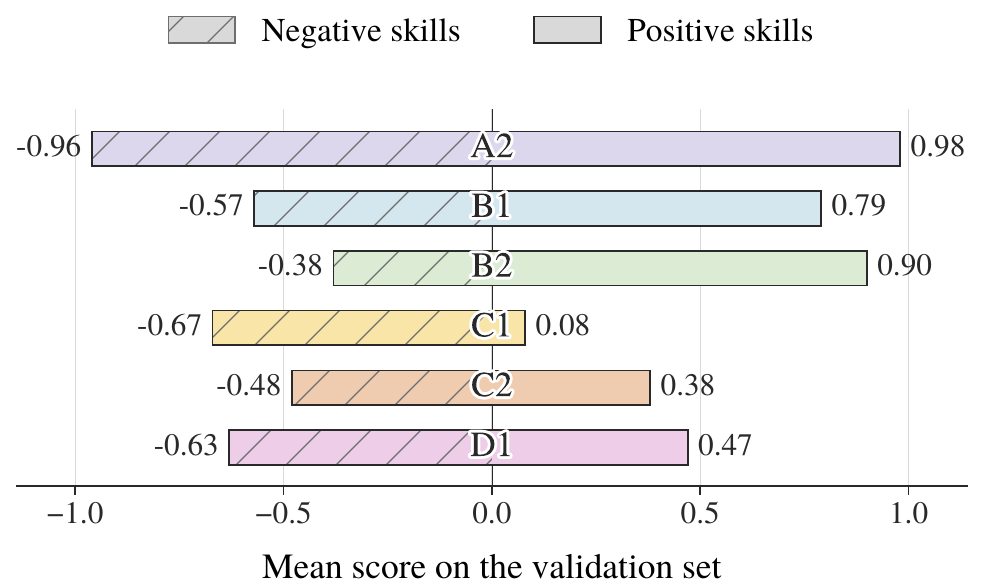}\\
 \caption{Mean validation scores for 259 positive and 259 negative skills across the six metrics (A2, B1, B2, C1, C2, and D1). Hatched bars extending to the left denote negative skills, whereas solid bars extending to the right denote positive skills. Values at the ends of the bars report the corresponding group means.\looseness=-1}
  \label{fig:positive_negative_bars}
\end{figure}

\noindent\textbf{Validation Score Distribution.}
We first examine whether the learned metric directions preserve the intended quality ordering on unseen skill documents. Each direction is constructed using only the training split and remains fixed during validation. We evaluate 518 held-out documents across A2, B1, B2, C1, C2, and D1, including 259 positive and 259 negative skills. A1 is excluded because it is evaluated using deterministic rules. As shown in Figure~\ref{fig:positive_negative_bars}, positive skills obtain mean scores above zero for all six metrics, whereas negative skills consistently receive scores below zero. Specifically, A2 shows the largest separation, with mean scores of $0.98$ and $-0.96$ for positive and negative skills, respectively. B1 and B2 also exhibit clear gaps of $1.36$ and $1.28$, followed by D1 with a gap of $1.10$. Although the separations are smaller for C2 and C1, their positive skills still outperform the corresponding negative skills by $0.86$ and $0.75$, respectively.
These results show that the learned directions recover the intended positive--negative ordering on held-out documents rather than merely memorizing individual training pairs. The variation in separation also provides insight into the properties captured by different directions. Explicit properties, such as trigger-condition clarity in A2, produce a sharper distinction, whereas workflow completeness and input--output clarity require information to be integrated across multiple parts of a document and therefore exhibit smaller gaps. This variation supports decomposing skill quality into metric-specific directions instead of representing it with a single holistic score. The resulting quality profile can distinguish skills of different quality and reveal which properties contribute most strongly to their differences.\looseness=-1

\begin{figure*}[t]
    \centering
    \includegraphics[width=1.0\textwidth]{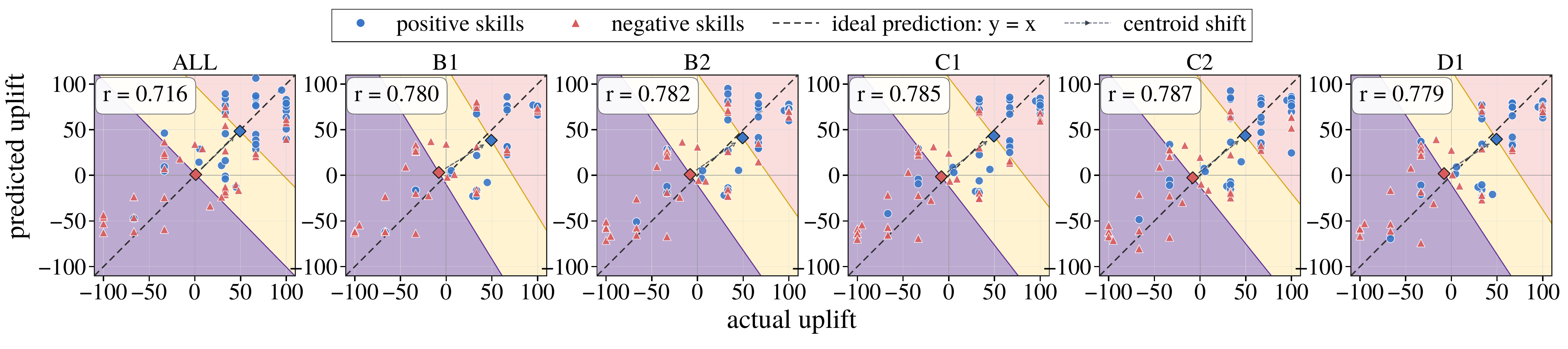}\\
   \caption{ Observed and predicted uplift for positive and negative skills, where uplift is the pass-rate difference between skills and no skills. The combined panel and the five metric-specific panels are partitioned into negative-skill, transition, and positive-skill regions along the centroid axis. Diamond markers denote the group centroids, the dashed arrow shows the centroid shift, and the dashed diagonal is the ideal reference line $y = x$.}
    \label{fig:single_metric_predicted}
\end{figure*}
\begin{figure}[t]
    \centering
    \includegraphics[width=1.0\columnwidth]{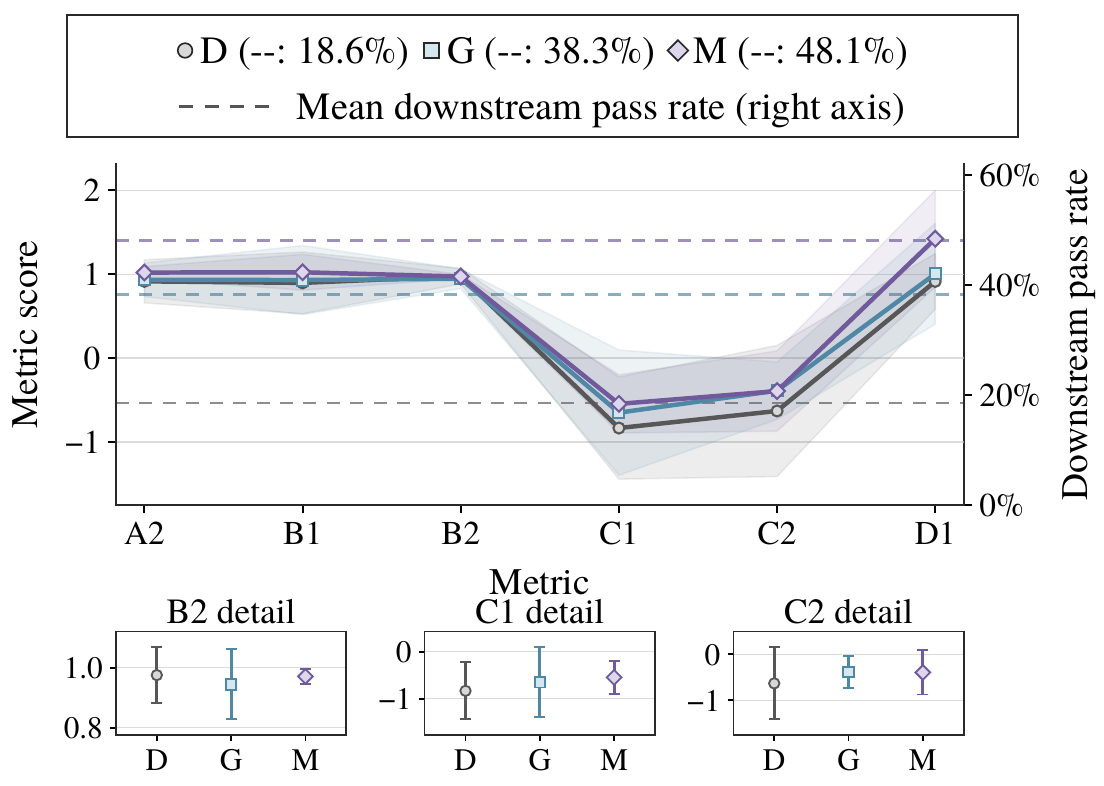}\\
   \caption{Comparison of metric profiles and downstream performance for skills before revision (D), LLM-only revisions (G), and metric-guided LLM revisions (M). Solid lines show the mean z-scored SkillEval scores, while horizontal dashed lines show the corresponding mean downstream pass rates on the right axis. The bottom panels provide enlarged views of B2, C1, and C2.}
    \label{d_g_best_m_metric_profiles}
\end{figure}
\noindent\textbf{Metric Correlation with Downstream Task Pass Rate.}
We use the \textsc{SkillsBench}~\citep{li2026skillsbenchbenchmarkingagentskills} examples with pass-rate uplift, defined as the difference between the pass rate with skills and the pass rate without skills, and fit one linear model per metric to map the metric score to uplift.  As shown in Figure~\ref{fig:single_metric_predicted}, all five metric-specific signals show positive correlations with downstream uplift. We do not include A2 in this analysis because A2 only evaluates the clarity of trigger conditions and is less directly related to the execution quality measured by downstream task performance. The correlations are consistently high across B1, B2, C1, C2, and D1, with Pearson $r$ ranging from 0.779 to 0.787, while the all-metric setting obtains $r=0.716$. The colored regions provide an additional distributional check: negative skills are concentrated more on the low-uplift side, whereas positive skills shift toward regions with larger uplift. This pattern indicates that the fitted linear functions preserve the expected positive--negative ordering of skills in the downstream task space. Together with the validation results in Figure~\ref{fig:positive_negative_bars}, this provides evidence that the learned metric scores are directionally aligned with downstream pass-rate variation.

\subsection{Skill Diagnosis and Targeted Revision}
We further explore whether SkillEval can diagnose weaknesses in skill documents and guide their revision. We select continuous-score tasks from \textsc{SkillsBench} for evaluation and compare three settings: skills before revision (D), LLM-only revisions that use D without access to SkillEval scores (G), and LLM revisions guided by the dimension-level diagnostics of SkillEval (M). 
As shown in Figure~\ref{d_g_best_m_metric_profiles}, LLM-only revision increases the mean downstream pass rate from 18.6\% to 38.3\%, while M further improves it to 48.1\%, corresponding to gains of 29.5 percentage points over D and 9.8 percentage points over G. The metric profiles show a similar pattern. Compared with D, M improves A2 from 0.914 to 1.019, B1 from 0.896 to 1.022, C1 from $-0.834$ to $-0.546$, C2 from $-0.631$ to $-0.393$, and D1 from 0.913 to 1.422, while maintaining a comparable B2 score. M also achieves higher scores than G on five of the six metrics, with C2 remaining nearly unchanged. These results indicate that SkillEval provides useful diagnostic feedback for skill revision. Guided by its metric-level scores, the LLM can focus revisions on weaker properties while retaining existing strengths, producing skills with improved quality profiles and higher downstream pass rates. \looseness=-1

\subsection{Additional Analyses}

\begin{figure}[t]
    \centering
    \includegraphics[width=1.0\columnwidth]{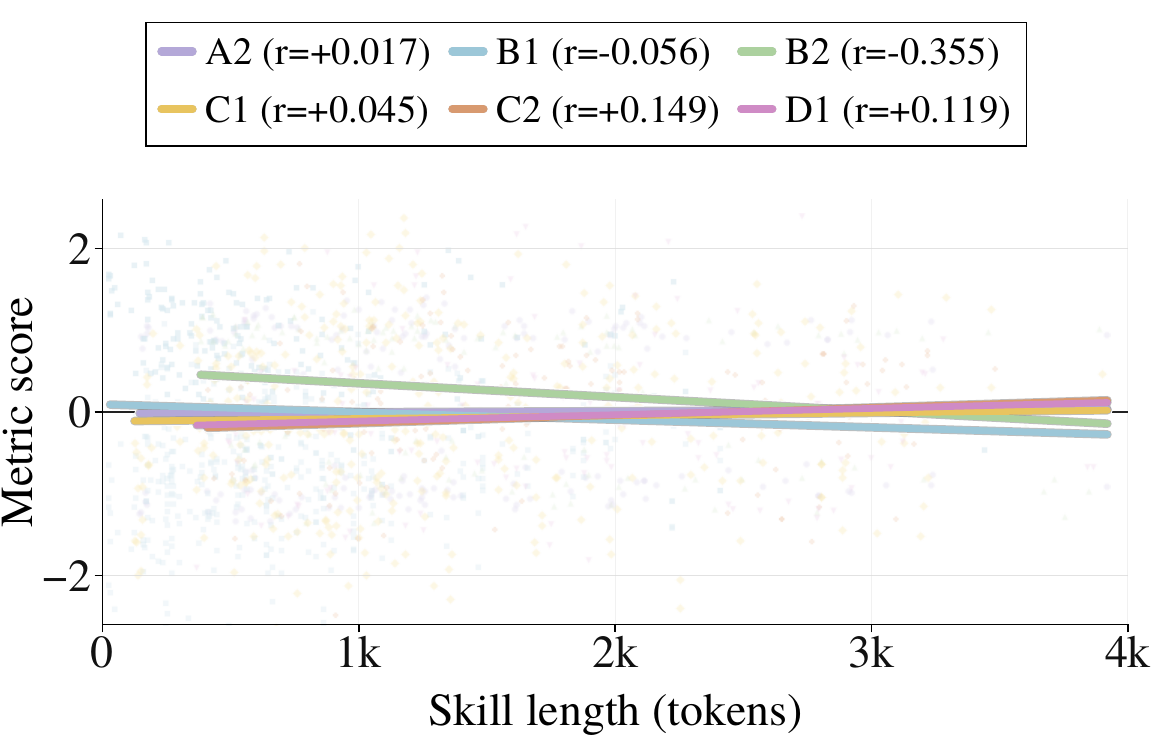}\\
   \caption{Each point represents an individual skill, with its token
  length on the horizontal axis and corresponding metric score on the vertical axis. Colored
  lines show the fitted linear trends, and the legend reports the Pearson correlation
  coefficient $r$ for each metric.}
    \label{fig:length_score_fit_linesn} 
\end{figure}
\noindent\textbf{Bias Analysis.}
We examine whether SkillEval scores are affected by the length of a skill document. For each semantic metric, we collect the set of skills scored by the frozen Qwen3.5-9B evaluator, represent each skill using its token length and standardized metric score. As shown in Figure~\ref{fig:length_score_fit_linesn}, most metrics show only weak correlations with length, including A2 ($r=0.017$), B1 ($r=-0.056$), B2($r=-0.355$), C1 ($r=0.045$), C2 ($r=0.149$), and D1 ($r=0.119$), indicating that the learned scores are not mainly determined by document length. Based on this observation, we reduce the influence of document length during scoring. Since the bias introduced by document length can be represented as a direction in the same hidden space as the metric directions, we orthogonalize each metric direction against a precomputed length direction before scoring. This reduces the direct influence of document length while preserving the intended semantic meaning of each metric. More bias removal experiments are provided in Appendix.

\begin{figure}[t]
    \centering
    \includegraphics[width=1.0\columnwidth]{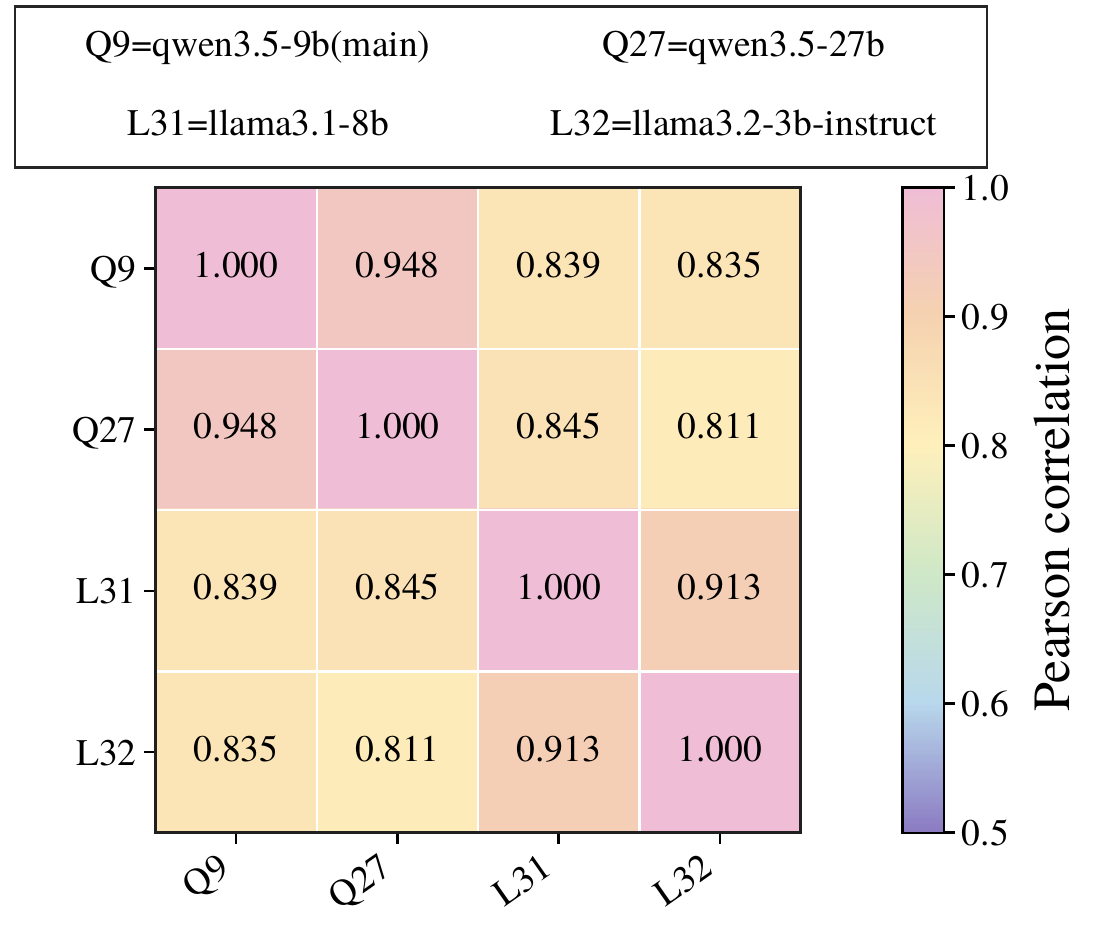}\\
   \caption{Analysis of the agreement between four scorer backbones (Q9, Q27, L31, and L32), based on the equal-weight mean Pearson correlation matrix of z-scored skill scores over the six metric scopes (A2, B1, B2, C1, C2, and D1). The
values summarize pairwise agreement across 1647 skills.\looseness=-1}
    \label{fig:test1_zscore_pearson_heatmap_mean}
\end{figure}

\noindent\textbf{Cross-Backbone Robustness of the Metrics.}
We evaluate scorer robustness by comparing four backbones, Qwen3.5-9B (Q9), Qwen3.5-27B (Q27), Llama3.1-8B (L31), and Llama3.2-3B-Instruct (L32), on six metric scopes (A2, B1, B2, C1, C2, and D1), over 1647 skills. For each scope, we compute the pairwise Pearson correlation matrix over z-scored metric outputs, and then average the six matrices with equal weight.
As shown in Figure~\ref{fig:test1_zscore_pearson_heatmap_mean}, the scorer outputs are highly consistent across backbones. The two Qwen models reach a Pearson correlation of 0.948, the two Llama models reach 0.913, and the cross-family pairs remain in the 0.811--0.845 range. This suggests that the scoring signal is largely preserved under backbone replacement, indicating a reasonable level of robustness in the scorer choice.

\begin{figure}[t]
    \centering
    \includegraphics[width=1.0\columnwidth]{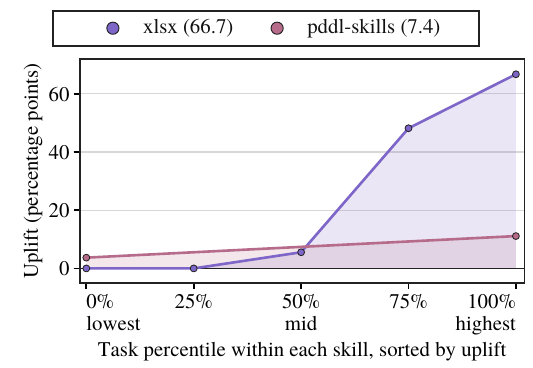}\\
   \caption{  Task-level pass-rate uplift of reused human-written skills. Tasks are sorted by uplift within each skill, and the legend reports the uplift gap between the lowest and highest task.\looseness=-1}
    \label{fig:single_skill_task_uplift}
\end{figure}
\noindent\textbf{Task Dependence of Downstream Skill Utility}
We analyze whether the downstream utility of a skill is stable across different task scenarios. In this experiment, we focus on reused human-written skills that appear in multiple downstream tasks with identical skill content. To avoid confounding from skill composition, we only consider tasks where the reused skill is the only available skill. For each skill--task pair, we compute the pass-rate uplift as the difference between the agent performance with the skill and without the skill.
Figure~\ref{fig:single_skill_task_uplift} presents the task-level uplift of reused skills, where tasks are sorted by uplift within each skill. We observe that the same skill can lead to substantially different improvements across tasks. Thus, identical skill content does not necessarily yield a consistent performance benefit across different task scenarios. For example, the reused \texttt{xlsx} skill produces almost no improvement on some tasks, but achieves a much larger uplift on tasks such as \texttt{weighted-gdp-calc} and \texttt{protein-expression-analysis}, with a maximum gap of 66.7 percentage points. In contrast, \texttt{pddl-skills} shows a smaller but still visible variation across its reused tasks, with a gap of 7.4 percentage points. The contrast between these examples further shows that the extent of task-dependent variation can differ across reused skills.
These results suggest that downstream performance is strongly affected by the match between a skill and the evaluated task. A single downstream task can show whether a skill is useful in that specific environment, but it cannot fully represent the overall quality of the skill across broader scenarios. This supports the need for complementary skill-level evaluation signals that assess reusable properties of the skill document.\looseness=-1

\section{Conclusion}
We introduced \textbf{SkillEval}, an interpretable framework for evaluating the document-level quality of agent skills. By decomposing skill quality into four complementary dimensions, SkillEval produces fine-grained quality profiles through an auditable and reproducible scoring process, while allowing potential biases to be identified and controlled. Our experiments show that these quality signals reliably distinguish skills of different quality and are strongly associated with their utility on downstream tasks. \looseness=-1

More broadly, our findings suggest that skill quality involves two complementary aspects: compatibility with a particular task and general properties that support reuse  across different scenarios. SkillEval connects these two aspects by providing interpretable document-level signals that are informative of downstream utility. We hope this work provides a foundation for more systematic skill evaluation and supports the development of skill ecosystems in which skills can be efficiently assessed, diagnosed, and refined. \looseness=-1

\section*{Limitations}

Our framework has several limitations and corresponding directions for future exploration. First, skill generation and targeted modification partly rely on LLMs. While the manual checks and controlled construction procedures described in Appendices~\ref{app:experimental_details} and~\ref{app:metric_dataset_construction} help ensure the intended attribute differences, the learned directions remain dependent on the quality and coverage of the underlying skills, and developing more robust skill collection and generation pipelines is a natural next step. Second, the semantic content of an attribute direction cannot be directly and fully characterized through quantitative measures alone. While the bias diagnostics and ablation studies in the appendices provide evidence that the learned directions capture their intended attributes, these analyses do not yield a complete quantitative account of their semantics or rule out all unidentified confounding factors. Extending the analysis through additional controlled interventions and independent semantic assessments would further characterize the interpretability and specificity of the learned directions.
\bibliography{custom}
\clearpage
\appendix

\begin{figure*}[t]
  \centering
  \includegraphics[width=\textwidth]{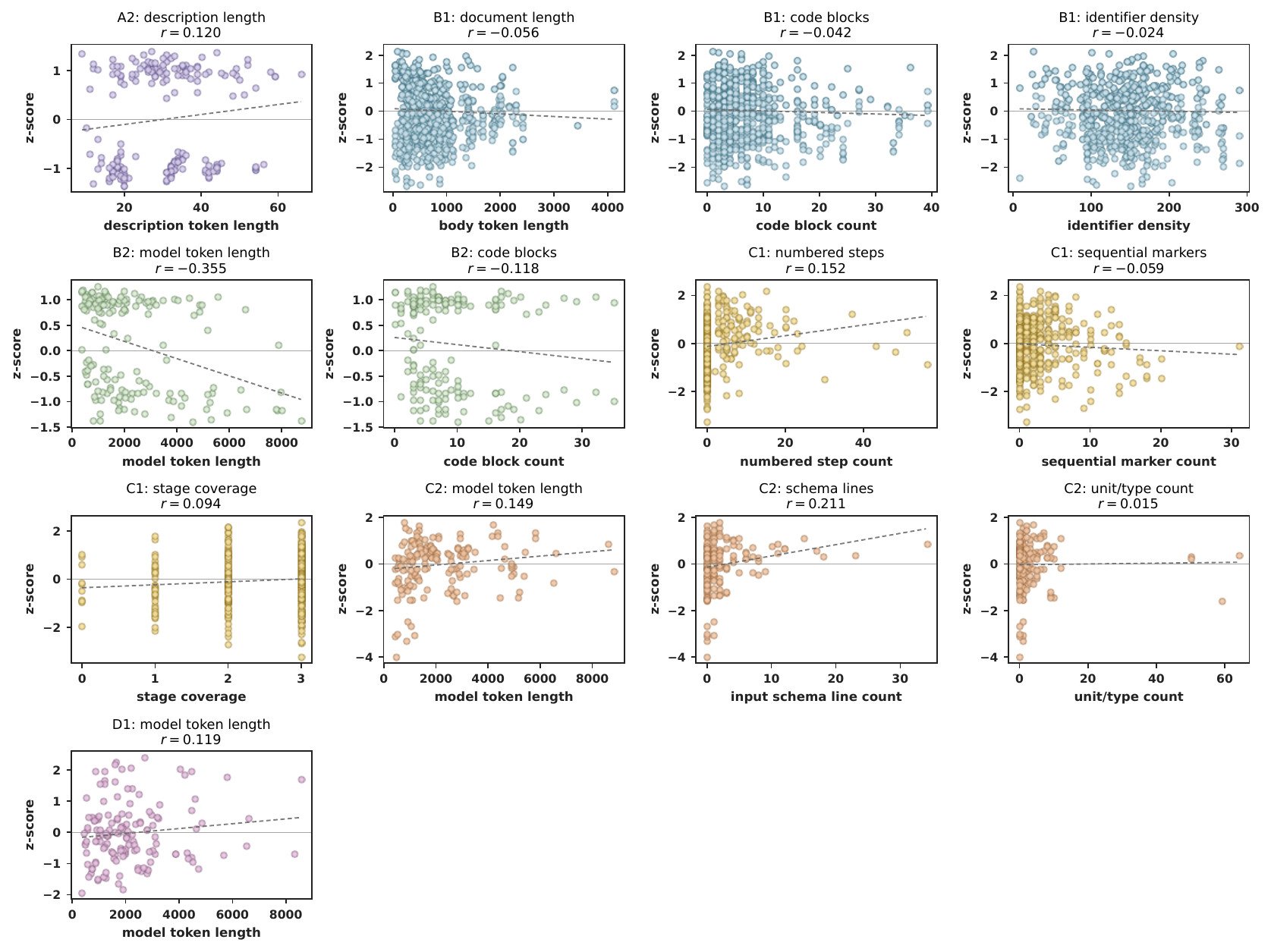}
  \caption{Bias diagnostics for SkillEval metrics on Qwen3.5-9B. Each panel shows the relationship between a selected bias variable and the
corresponding z-scored metric output. }
  \label{fig:bias_diagnostic}
\end{figure*}

\section{Experimental Details}
\label{app:experimental_details}
We used separate training and validation sets in all
experiments. For experiments involving scoring-direction
estimation or linear-model fitting, parameters were
estimated exclusively from the training set, and the
reported performance and correlations were evaluated
on the validation set. For binary-outcome tasks, pass rate denotes the fraction of successful runs among three trials, whereas for continuous-score tasks, we report the mean benchmark score over three trials on a 0--100 scale;

For each evaluation attribute, both positive and negative
skills underwent human and AI review. The review checked
the intended quality difference in the target attribute
and consistency in the remaining attributes. We retained
only pairs that differed in the target attribute while
preserving all other content and attributes, thereby
reducing the influence of non-target factors.

In the supplementary experiments, Qwen3.5-9B ran in
evaluation mode with BF16 precision, and the extracted
hidden vectors were stored in FP32. The software environment
used PyTorch 2.9.1, Transformers 5.5.4, and CUDA 12.8.

For accuracy comparisons, macro scores were computed
as unweighted averages across the six semantic metrics.
Pairwise accuracy was first computed within each skill
family and then averaged equally across families, preventing
families with more negative variants from dominating
the results. We estimated 95\% confidence intervals using
2,000 bootstrap resamples of validation task groups,
with the fitted models held fixed. All compared methods
shared the same resampling draws to ensure consistent
comparisons.

\section{Metric-Specific Dataset Construction}
\label{app:metric_dataset_construction}

\paragraph{Data Sources and Common Procedure.}
We construct a separate dataset for each of the six learned semantic metrics using three complementary sources. First,
LLMs generate complete positive and negative \texttt{SKILL.md} documents according to metric-specific instructions.
Second, we collect publicly available skill documents from online sources and screen them as positive or negative
examples based on the corresponding metric criteria. Therefore, the negative examples are not limited to controlled
rewrites of positive skills: they also include LLM-generated documents and collected documents that naturally exhibit
the weakness measured by a metric. Third, we construct controlled positive--negative pairs using metric-specific
operations when a clearer comparison is needed. The same criteria are used to filter generated, collected, and
constructed examples. During controlled construction, we preserve the task goal, topic, tools, document structure, and
non-target properties as much as possible, so that the main difference within each pair is the property evaluated by
the corresponding metric. All examples retain the data split of their source task to prevent related skills from
appearing in both training and validation sets. A1 is excluded because Structural Format Validity is evaluated directly
using deterministic rules.

\subsection{Format Validity}
A1 evaluates whether a skill document satisfies the minimum structural requirements using deterministic, model-independent rules. Specifically, we check whether the file has a \texttt{.md} extension, begins with YAML frontmatter enclosed by \texttt{---} delimiters, and contains both the \texttt{name} and \texttt{description} keys in a successfully parsed YAML mapping. A document passes A1 only if all checks are satisfied; otherwise, the checker returns the corresponding failure reasons. These checks assess field presence without imposing constraints on field values. A1 thus provides a binary assessment of structural validity and requires neither hidden-state extraction nor direction learning, while the remaining six metrics assess semantic quality.

\subsection{Trigger Clarity}

The A2 dataset contains LLM-generated and online-collected skills whose descriptions either clearly specify or fail to
specify when the skill should be used. Positive examples state an explicit trigger condition and connect it to a
concrete task scenario, whereas negative examples provide vague conditions or only summarize the tool's general
capability. For controlled pairs, an LLM rewrites only the frontmatter description, replacing a clear trigger condition
with a general tool or capability description. The skill body and all other frontmatter fields remain unchanged. We
also keep the positive and negative descriptions similar in length. Generated, collected, and rewritten examples are
screened using the same trigger-expression and task-scenario criteria.

\subsection{Content Consistency}

The B1 dataset combines generated and collected skills with either consistent or inconsistent descriptions and bodies.
Positive examples describe the same task in the frontmatter and body, whereas negative examples contain a description
that does not match the procedural content. To construct additional controlled negatives, we preserve the original
frontmatter while replacing the body with one taken from a different skill. Replacement bodies are matched to the
original body in length where possible. This operation changes description--body consistency while preserving the
description, metadata format, and approximate document scale. We record the source of each replacement body and check
that the frontmatter remains unchanged and that the paired bodies are different.

\subsection{Technical Specificity}

The B2 dataset includes generated and collected skills with different levels of technical specificity. Positive
examples contain task-related technical details, such as API names, function calls, parameters, configuration values,
file names, schema fields, numerical constants, units, and usable code or command examples. Negative examples include
LLM-generated or collected skills that provide only general technical guidance. For controlled pairs, an LLM rewrites a
specific skill by replacing concrete technical details with more general descriptions while preserving the frontmatter,
task topic, main workflow, input--output setting, and overall structure. The rewritten document is kept close to the
original in length. All candidates are screened using the amount and density of task-related technical evidence.

\subsection{Workflow Integrity}

The C1 dataset contains generated and collected skills with complete or incomplete workflows. Positive examples provide
an ordered procedure that connects the input, intermediate operations, and final output. Negative examples include
skills with missing stages, disordered steps, weak transitions, or disconnected recommendations. We construct
additional controlled negatives by asking an LLM to reorder dependent steps, remove an important transition, replace an
ordered procedure with unordered suggestions, or make the path from input to output incomplete. The task, tools,
input--output setting, and most technical details are retained. The resulting examples are screened to ensure that
workflow integrity is reduced without substantially changing the task content or other quality properties.

\subsection{I/O Explicitness}

The C2 dataset combines generated and collected skills with different levels of input--output explicitness. Positive
examples specify inputs and outputs using concrete formats, fields, columns, keys, schemas, units, data types,
constraints, or output labels. Negative examples include generated and collected documents in which this information is
missing or expressed only in general terms. For controlled construction, we preserve the frontmatter, task, tools,
APIs, and main workflow while removing or generalizing input--output details. Explicit descriptions such as CSV fields,
JSON keys, units, types, and output schemas are replaced with general references to input data or result summaries.
Content unrelated to the input--output contract is kept unchanged, and large deletions are avoided to control document
length.

\subsection{Failure Awareness}

The D1 dataset contains LLM-generated and online-collected skills with different levels of failure awareness. Positive
examples describe potential failures, boundary conditions, incorrect practices, diagnostic signals, or recovery
strategies. Negative examples include generated and collected skills that omit such guidance or mention it only
superficially. For controlled pairs, rule-based processing first removes complete failure-related sections and
standalone warnings. An LLM then performs minimal edits to failure-related clauses embedded in otherwise useful
sentences. The frontmatter and unrelated procedural content are preserved. The resulting examples are screened for
failure-related sections, explicit prohibitions, descriptions of common mistakes, explanations of failure causes, and
corrective guidance.

\section{Bias Diagnostic}

We further examine whether SkillEval scores are affected by common biases that are not part of the intended semantic meaning of each
metric. In this experiment, we select representative bias variables for each metric and compute their Pearson correlations with the
corresponding z-scored metric outputs.

Specifically, for A2, we test description length, defined as the token length of the trigger description, since A2 is evaluated from
the trigger description. For B1, we test document length, code block count, and identifier density, where document length is measured
by the token length of the skill body, code block count measures the number of Markdown code blocks, and identifier density measures
the density of technical identifiers such as API names, file names, field names, and numeric expressions. These variables may reflect
document scale or formatting rather than description-body alignment.

For B2, we test model token length and code block count, since specificity scores may be affected by longer or more code-heavy skill
documents. For C1, we test numbered step count, sequential marker count, and stage coverage. Numbered step count measures explicit
ordered steps, sequential marker count measures temporal markers such as ``first'', ``then'', and ``finally'', and stage coverage
measures whether the skill covers input, processing, and output stages. These variables correspond to common workflow-format cues.

For C2, we test model token length, schema-line count, and unit/type count, where schema-line count measures lines describing input
fields, columns, keys, units, types, or constraints, and unit/type count measures explicit unit or data-type expressions. These
variables may affect input-output explicitness. For D1, we test model token length to examine whether failure-awareness scores are
mainly driven by skill length.

As shown in Figure~\ref{fig:bias_diagnostic}, most selected bias variables show weak correlations with the corresponding metric
scores. A2 has only a weak correlation with description length ($r=0.120$). B1 is weakly correlated with document length
($r=-0.056$), code block count ($r=-0.042$), and identifier density ($r=-0.024$). B2 shows a moderate negative correlation with
model token length ($r=-0.355$) and a weak correlation with code block count ($r=-0.118$). C1 is weakly correlated with numbered
step count ($r=0.152$), sequential marker count ($r=-0.059$), and stage coverage ($r=0.094$). C2 also shows weak correlations with
model token length ($r=0.149$), schema-line count ($r=0.211$), and unit/type count ($r=0.015$). For D1, the correlation between
model token length and the metric score is also weak ($r=0.119$).

These results suggest that most SkillEval scores are not mainly determined by the selected surface-level biases. At the same time,
the diagnostic still reveals metric-specific sensitivities, such as the moderate length effect observed for B2. This provides a
practical audit mechanism for SkillEval: after a metric direction is learned, we can test whether its scores align with intended
semantic properties or with unintended document cues. When a bias shows a visible correlation, it can be further controlled during
metric construction or removed by orthogonalizing the metric direction against the corresponding bias direction in hidden space.

\section{Comparison with Structural and Length Baselines}
\label{sec:appendix_surface_baselines}

We compared SkillEval with structural and length-based predictors on the same validation set across six semantic metrics: A2, B1, B2, C1, C2, and D1. SE-RAW used the original mean-difference scoring directions, whereas SE-DEB removed their components along a precomputed length direction. Both variants used frozen Qwen3.5-9B representations from hidden-state index 16, retaining the metric-specific prefixes and pooling rules. The structural baseline (STR) used 61 fixed features describing document structure and procedural markers, while the length baseline (LEN) used only the log-transformed token count. Both baselines used L2-regularized logistic regression, with regularization selected through five-fold task-group cross-validation within the training set. We evaluated document-level AUROC and positive--negative pair accuracy (PairAcc), assigning ties a score of 0.5 and averaging pair accuracy within skill families before aggregation. Macro scores were averaged equally across the six metrics, with 95\% confidence intervals estimated from 2,000 paired validation-task bootstrap resamples.

\begin{figure*}[t]
  \centering
  \includegraphics[width=\textwidth]{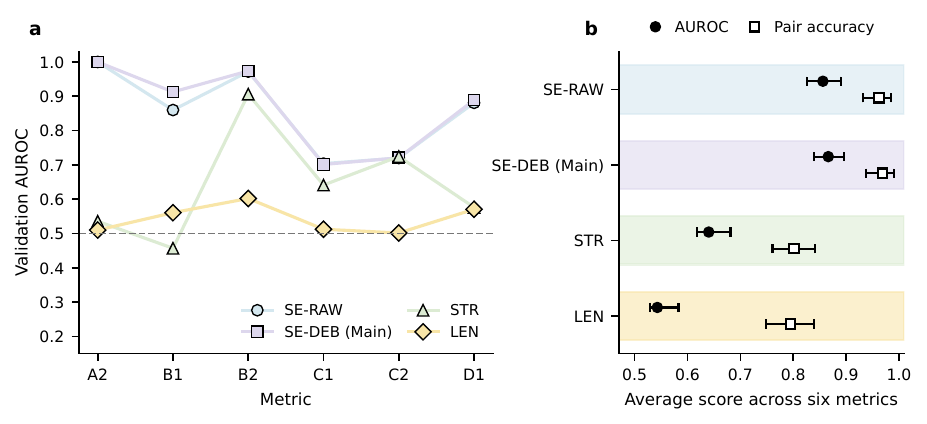}
  \caption{Validation performance of SkillEval and structural and length-based baselines. SE-RAW and SE-DEB denote the original and length-debiased SkillEval variants, respectively. STR and LEN use structural features and log-transformed token counts, respectively. (a) Per-metric AUROC across A2, B1, B2, C1, C2, and D1; the dashed line indicates chance-level performance (0.5). (b) Macro-averaged AUROC (filled circles) and positive--negative pair accuracy (open squares), with equal weighting across the six metrics. Error bars show 95\% confidence intervals estimated from 2,000 paired bootstrap resamples of validation tasks.}
  \label{fig:appendix_surface_baselines}
\end{figure*}

Figure~\ref{fig:appendix_surface_baselines} presents the per-metric AUROC and the aggregate results. SE-RAW achieved a Macro-AUROC of 0.856 (95\% CI: 0.825--0.890), while SE-DEB achieved 0.866 (0.840--0.896). Their Macro-PairAcc values were 0.962 and 0.969, respectively. In comparison, STR achieved a Macro-AUROC of 0.640 (0.618--0.681), and LEN achieved 0.543 (0.530--0.583). The paired Macro-AUROC difference between SE-DEB and SE-RAW was 0.010 (0.001--0.020), indicating a modest aggregate improvement after removing the precomputed length component.

These results indicate that SkillEval provides stronger aggregate discrimination than the tested structural and length predictors on this constructed evaluation set. The advantage varies across metrics, so the macro-average should be interpreted alongside the per-metric results. LEN also achieved a relatively high Macro-PairAcc of 0.795 despite its near-chance Macro-AUROC. This distinction shows that correctly ordering matched positive--negative pairs does not necessarily imply consistent discrimination across different skill families, motivating the joint reporting of both measures.

\section{Effect of Training Label Randomization}
\label{sec:appendix_random_labels}

We randomized the training labels to assess the contribution of correct quality supervision to SkillEval's discrimination performance. We held the training--validation split, hidden representations, and direction estimation procedure fixed, varying only the training labels. In each repeat, we independently retained or flipped the positive--negative labels at the task level, applied the same transformation to all skills and their negative variants within that task, and re-estimated the scoring directions. Validation labels remained unchanged, and direction signs were not adjusted based on validation performance. We performed 500 randomization runs and included all results.

\begin{figure*}[t]
  \centering
  \includegraphics[width=\textwidth]{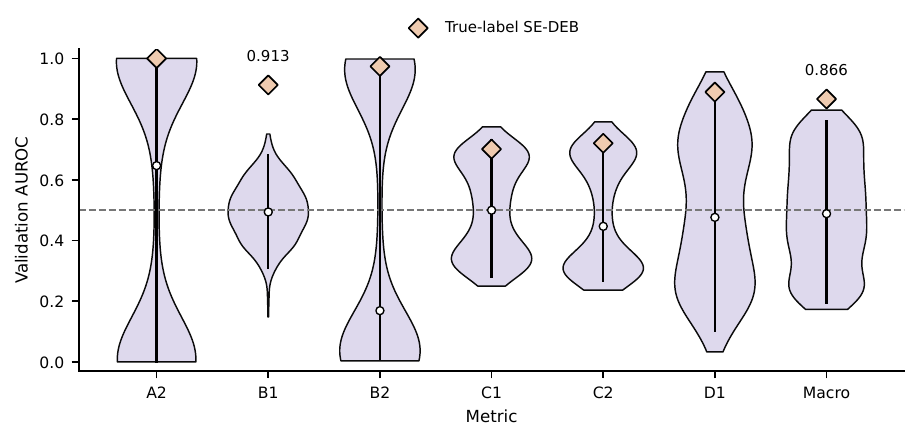}
  \caption{
Effect of training label randomization.
Violins show the smoothed distributions of validation AUROC
across all 500 RAND-SE fits on the fixed validation set.
White circles indicate medians, and black vertical lines
span the 2.5th--97.5th percentiles of these randomized-fit
results. These ranges describe variability across randomized
training-label assignments; 
Diamonds indicate SE-DEB performance with the original
training labels, and the horizontal dashed line marks
chance-level AUROC (0.5).
Macro denotes the unweighted mean AUROC across the six metrics.
}
  \label{fig:appendix_random_labels}
\end{figure*}

Figure~\ref{fig:appendix_random_labels} compares the AUROC distributions of RAND-SE with SE-DEB trained on the original labels. RAND-SE achieved a mean Macro-AUROC of 0.492, with a standard deviation of 0.187 and a 2.5th--97.5th percentile range of 0.196--0.795. SE-DEB achieved a Macro-AUROC of 0.866, exceeding the upper bound of this range. Sensitivity to label randomization varied across metrics: A2 and B2 exhibited broad distributions, with some randomized directions still achieving high AUROC. These percentile ranges describe variation across randomized training labels rather than confidence intervals for a fixed model.

Label randomization reduced average aggregate performance to approximately chance level, indicating that correct quality labels provide informative supervision for estimating the scoring directions.

\section{Cross-Model Robustness of Skill Utility}
We further evaluate whether the downstream utility of skills is stable across different execution models. The evaluation data are
provided by \textsc{SkillsBench}, which contains downstream task results under both with-skill and no-skill conditions. Following the
\textsc{SkillsBench} setting, we use results from 87 tasks evaluated with 18 downstream executor models.

For each task and executor model, we compute the raw uplift as the pass-rate difference between the with-skill and no-skill
conditions. Since each condition is evaluated with three trials, the smallest pass-rate resolution is $1/3$. We therefore use a
tolerance-aware pairwise concordance metric with margin $0.333$: for each pair of executor models, we compare task pairs only when the
uplift gap is at least this margin for both models, and count whether the two models rank the task pair in the same order.

\begin{figure*}[t]
  \centering
  \includegraphics[width=\textwidth]{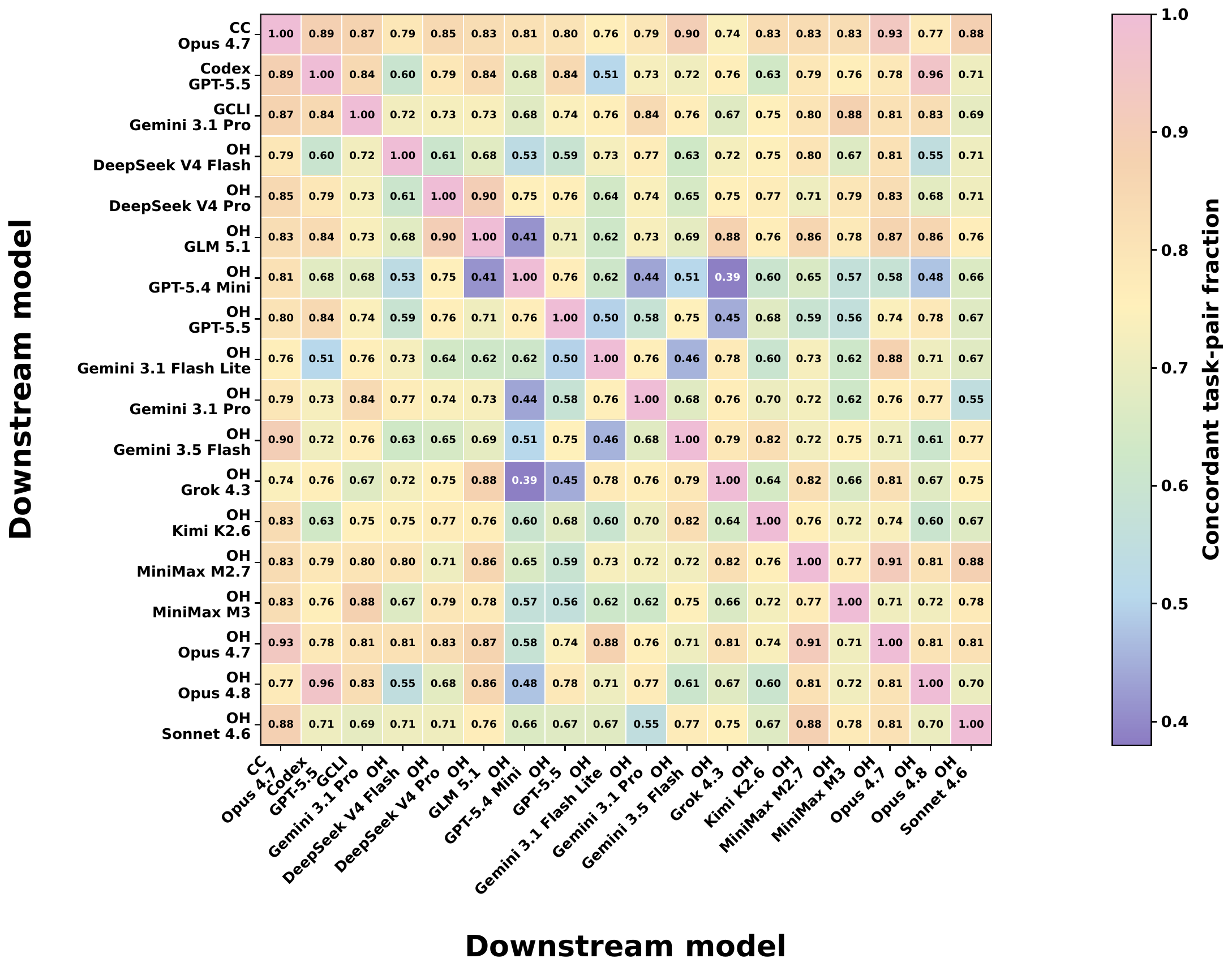}
  \caption{  Heatmap of tolerance-aware pairwise concordance between downstream executor models using \textsc{SkillsBench} evaluation results. Rows and columns denote executor models, and each cell reports the fraction of task pairs for which two models produce the same relative
ordering of raw skill uplift. }
  \label{fig:model_tolerance_pairwise_concordance}
\end{figure*}

As shown in Figure~\ref{fig:model_tolerance_pairwise_concordance}, the average concordance is $0.724$ and the median concordance is $0.739$. In total, 103 out of 153
model pairs have concordance above $0.7$, and 146 out of 153 model pairs are above $0.5$. Several model pairs show particularly
strong agreement, such as Codex/GPT-5.5 and OpenHands/Opus 4.8 ($0.956$), and Claude Code/Opus 4.7 and OpenHands/Opus 4.7
($0.931$). At the same time, a small number of model pairs have lower agreement, with the minimum concordance observed between
OpenHands/GPT-5.4 Mini and OpenHands/Grok 4.3 ($0.385$).

These results indicate that skill-induced uplift is largely consistent across different downstream executor models, suggesting that
the measured utility of skills is not tied to a single execution model. However, the lower-concordance cases also show that executor
choice can still affect the relative ordering of skill benefits. Therefore, downstream skill evaluation is reasonably robust across
models, but cross-model analysis remains useful for identifying model-specific sensitivity.

\section{Hidden-State Layer Analysis}
\label{app:layer_analysis}

\paragraph{Experimental Setup.}
We examine how the choice of hidden-state layer affects the metric directions learned by SkillEval. Using the frozen
Qwen3.5-9B model, we sweep hidden-state indices from 4 to 32 at intervals of four. For each metric, we use the same
metric-specific training pairs across layers and retain the pooling rule used in the main experiments. 

For metric $i$ and hidden-state index $\ell$, we project the pooled representation of a skill $d$ onto the
corresponding metric direction and standardize the resulting score using the training-split statistics:
\[
z_i^{(\ell)}(d)
=
\frac{s_i^{(\ell)}(d)-\mu_i^{(\ell)}}
{\sigma_i^{(\ell)}}.
\]
We evaluate each layer using two complementary measures. Validation AUROC measures how well the standardized scores
distinguish positive skills from negative skills. Specifically, we fit a one-dimensional logistic classifier to the
standardized training scores and compute AUROC from its predicted probabilities on the corresponding validation split.
The mean paired margin measures the average standardized score difference within matched positive--negative validation
pairs:
\[
M_i^{(\ell)}
=
\frac{1}{N_i}
\sum_{j=1}^{N_i}
\left[
z_i^{(\ell)}(d_{i,j}^{+})
-
z_i^{(\ell)}(d_{i,j}^{-})
\right],
\]
where $N_i$ is the number of validation pairs for metric $i$. A larger AUROC indicates better discrimination between
positive and negative skills, while a larger paired margin indicates stronger separation within matched pairs.

\subsection{Validation AUROC}
\label{app:layer_auc}

\begin{figure*}[t]
    \centering
    \includegraphics[width=\textwidth]{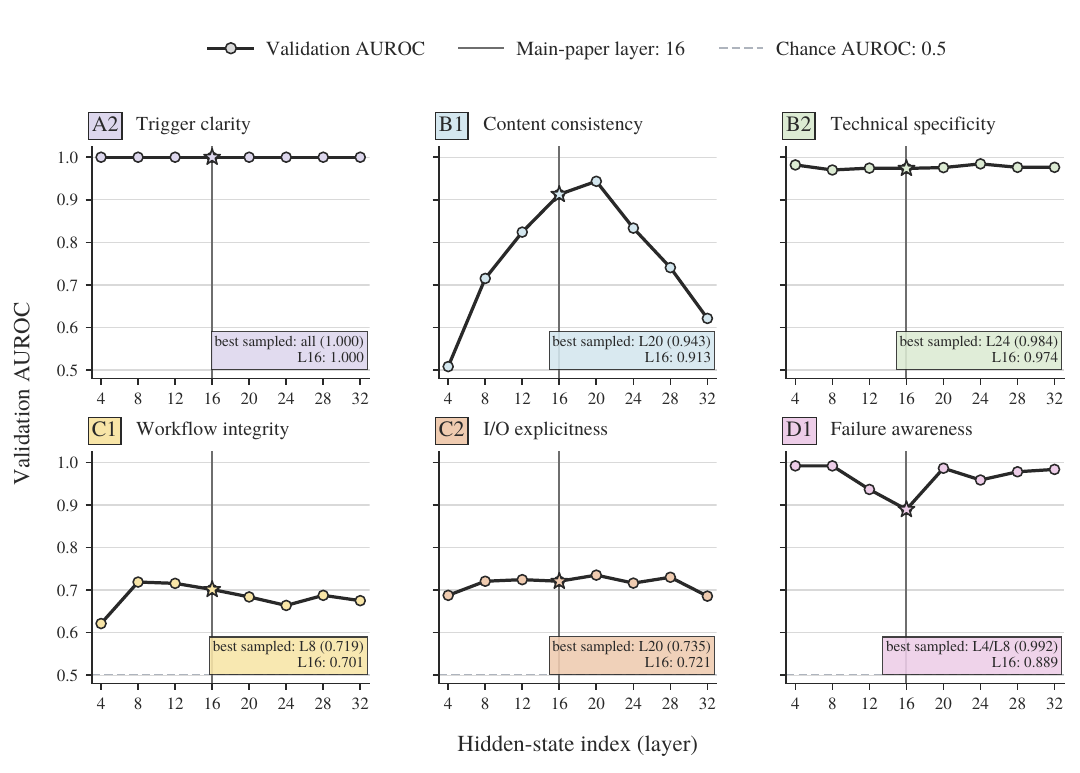}
    \caption{Validation AUROC across hidden-state indices for the six learned SkillEval metrics. AUROC is computed on
    the corresponding validation split using predicted probabilities from a one-dimensional logistic classifier fitted
    to the standardized training scores. The solid vertical line and star indicate hidden-state index 16 used in the
    main experiments, and the dashed horizontal line denotes chance-level AUROC. Each annotation reports the best
    sampled layer and its AUROC together with the AUROC at layer 16.}
    \label{fig:layer_auc}
\end{figure*}

Figure~\ref{fig:layer_auc} compares the validation AUROC obtained from different hidden-state indices. A2 achieves an
AUROC of 1.000 at every sampled layer, while B2 remains stable between 0.970 and 0.984. B1 is more sensitive to layer choice: its AUROC increases from 0.508 at layer 4 to 0.913 at layer 16 and reaches its maximum of 0.943 at layer 20. At
layer 16, C1 and C2 achieve AUROCs of 0.701 and 0.721, respectively, compared with their sampled maxima of 0.719 at
layer 8 and 0.735 at layer 20. D1 obtains an AUROC of 0.889 at layer 16, while its highest value of 0.992 occurs at
layers 4 and 8.

For A2, B1, B2, C1, and C2, the AUROC at layer 16 is within 0.031 of the best sampled result. D1 exhibits greater
variation across layers, although its AUROC at layer 16 remains substantially above the chance level of 0.5. These
results indicate that the layer producing the strongest discrimination varies across metrics, while layer 16 retains
reliable discrimination for all six metrics.

\subsection{Mean Paired Margin}
\label{app:layer_margin}

\begin{figure*}[t]
    \centering
    \includegraphics[width=\textwidth]{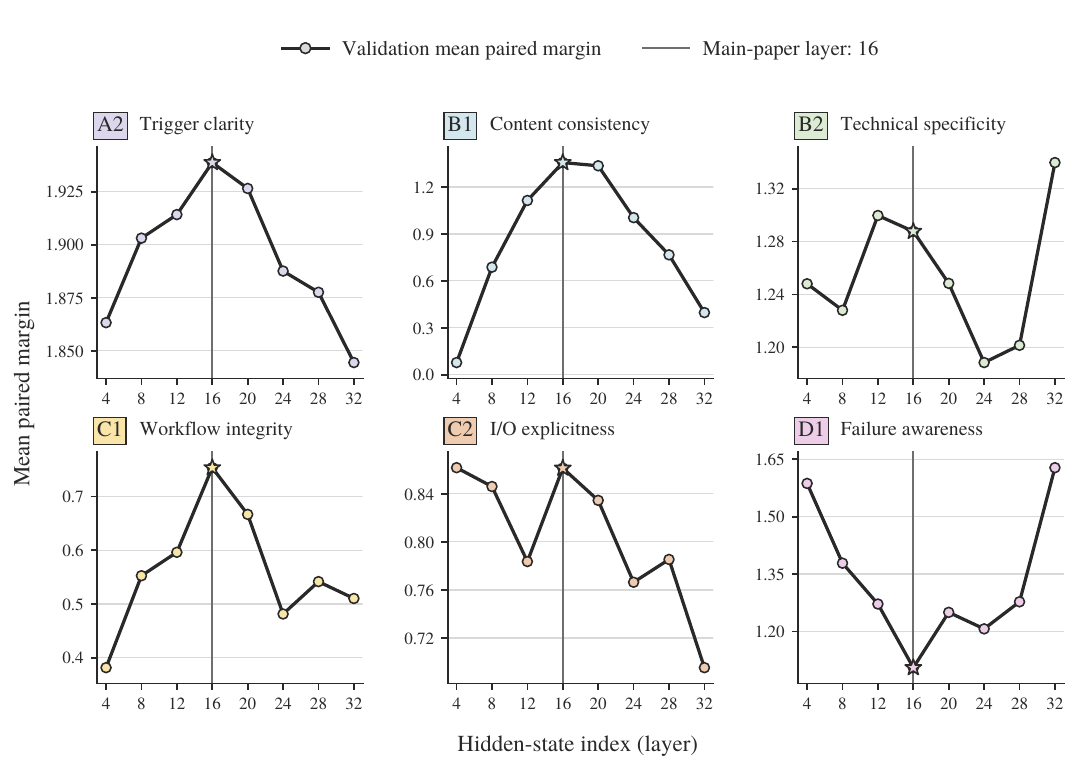}
    \caption{Validation mean paired margin across hidden-state indices for the six learned SkillEval metrics. The
    vertical axis reports the mean difference between the standardized scores of matched positive and negative
    validation skills. The solid vertical line and star indicate hidden-state index 16 used in the main experiments.}
    \label{fig:layer_margin}
\end{figure*}

We further evaluate each layer using the mean paired margin. As shown in Figure~\ref{fig:layer_margin}, all six metrics
retain positive margins at layer 16, indicating that positive skills receive higher scores than their matched negative
counterparts on average. The margins at layer 16 are 1.939 for A2, 1.356 for B1, 1.288 for B2, 0.754 for C1, 0.861 for
C2, and 1.105 for D1.

Layer 16 produces the largest sampled margins for A2, B1, and C1. For C2, its margin at layer 16 differs from the
sampled maximum at layer 4 by less than 0.001. B2 and D1 obtain their largest margins at layer 32, reaching 1.340 and
1.628, respectively. The layers that maximize the paired margin therefore do not always coincide with those that
maximize AUROC, showing that score separation and ranking performance provide complementary views of each metric
direction.

Taken together, these results show that no single hidden-state layer is optimal for every metric.

\section{Score Consistency and Robustness to Formatting}
\label{sec:appendix_stability}


\begin{table*}[t]
\centering

\begin{tabular}{@{}lrrrrr@{}}
\toprule
\multicolumn{6}{@{}l}{
    \textbf{(a) Repeated-input consistency}
} \\
\midrule
Method & $n$ & Repeats & Max.\ SD
       & Max.\ $|\Delta s|$ & Max.\ flip rate \\
\midrule
SE-RAW & 258 & 5 & 0 & 0 & 0.000 \\
SE-DEB & 258 & 5 & 0 & 0 & 0.000 \\
\bottomrule
\end{tabular}

\medskip

\begin{tabular}{@{}llrrcr@{}}
\toprule
\multicolumn{6}{@{}l}{
    \textbf{(b) Robustness to formatting}
} \\
\midrule
Method & Format & $n$ & Min.\ $\rho$
       & $\Delta$AUROC range & Max.\ flip rate \\
\midrule
SE-RAW & F1 & 256 & 0.982
       & $[-0.0250,\; 0.0100]$ & 0.050 \\
SE-RAW & F2 & 253 & 0.993
       & $[ 0.0000,\; 0.0075]$ & 0.000 \\
SE-DEB & F1 & 256 & 0.977
       & $[-0.0375,\; 0.0025]$ & 0.050 \\
SE-DEB & F2 & 253 & 0.990
       & $[ 0.0000,\; 0.0079]$ & 0.000 \\
\bottomrule
\end{tabular}

\caption{
Score consistency and robustness to formatting.
$n$ denotes the number of metric--document entries.
(a) Each entry undergoes five independent representation
extractions and scoring runs.
SD and $|\Delta s|$ measure score variation across repeats.
(b) F1 modifies blank lines; F2 replaces equivalent list
markers or heading forms.
$\rho$ is the Spearman correlation between original and
transformed scores, and $\Delta$AUROC is the transformed
AUROC minus the original AUROC on the same accepted entries.
The minimum correlation, AUROC-change range, and maximum
pairwise ranking flip rate are computed across the six
metrics. The ranges are not confidence intervals.
}
\label{tab:appendix_stability}
\end{table*}

We evaluated SkillEval's stability through repeated scoring
and content-preserving formatting changes. For repeated
scoring, we performed five independent representation
extractions and scoring runs for 258 metric--document
entries, keeping the model, scoring directions, and execution
environment fixed. For formatting changes, we constructed
two variants: F1 modifies blank lines outside YAML and code
blocks, while F2 replaces equivalent list markers or uses
an equivalent heading form. Each transformation was checked
to ensure that the YAML content and parsed Markdown content
remained unchanged. Of the 516 candidate transformations,
509 passed these checks, comprising 256 F1 entries and
253 F2 entries. Original and transformed documents were
evaluated on the same accepted entries.

Table~\ref{tab:appendix_stability} summarizes the results.
Both SE-RAW and SE-DEB produced identical scores across
all five repeated runs, with zero maximum score standard
deviation, zero maximum absolute score change, and no
pairwise ranking flips. Across the six metrics and two
formatting variants, SE-DEB achieved a minimum Spearman
rank correlation of 0.977, with AUROC changes ranging
from $-0.0375$ to $0.0079$ and a maximum pairwise ranking
flip rate of 5\%. The corresponding results for SE-RAW
were 0.982, $-0.0250$ to $0.0100$, and 5\%, respectively.
The AUROC ranges report the minimum and maximum changes
across metrics rather than confidence intervals.

These results indicate that SkillEval produces reproducible
scores under fixed execution conditions and preserves
high ranking consistency under the two tested formatting
changes, providing evidence of its robustness to these
surface variations.

\section{Pooling Strategy Analysis}
\label{app:pooling_analysis}

\paragraph{Experimental Setup.}
We further examine how the pooling strategy affects the metric directions learned by SkillEval. At hidden-state index
16, we compare two general pooling strategies while keeping the model, prompts, skill pairs, and evaluation procedure
fixed. Full-mean pooling averages the hidden representations of all tokens in the complete input sequence, whereas
last-token pooling uses only the representation of the final token. For each pooling strategy, we reconstruct the
metric direction from the same metric-specific training pairs and evaluate it on the corresponding fixed validation
split. For the last-token configuration, we re-estimate the length direction using last-token representations.

\begin{figure*}[t]
    \centering
    \includegraphics[width=\textwidth]{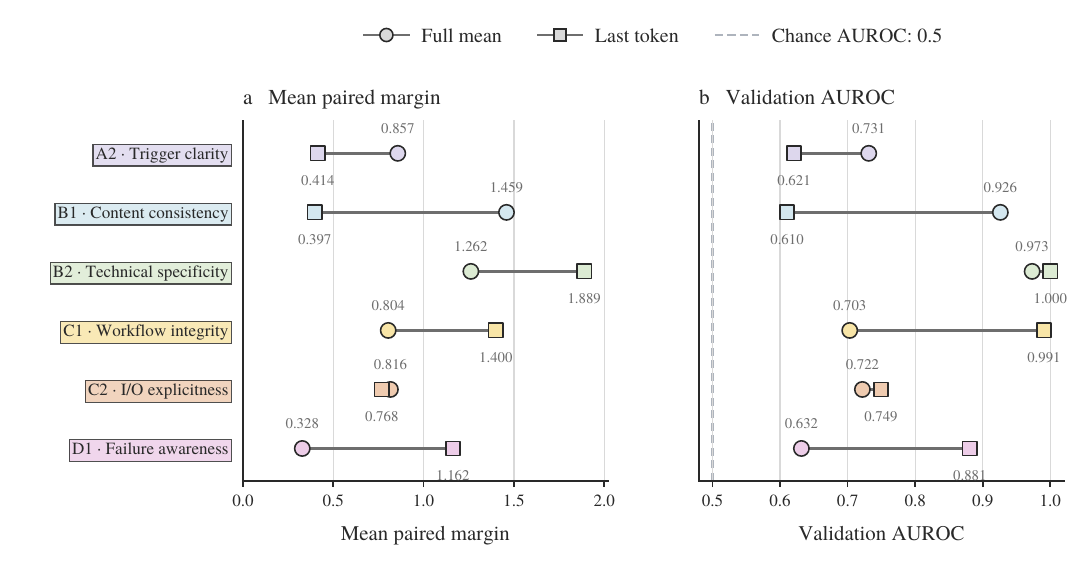}
    \caption{Comparison of full-mean and last-token pooling at hidden-state index 16. Figure (a) shows the mean difference
in standardized scores between matched positive and negative validation skills. Figure (b) shows the validation AUROC.
Circles show full-mean pooling, squares show last-token pooling, and the dashed line marks the chance AUROC of 0.5.}
    \label{fig:pooling_analysis}
\end{figure*}

The mean paired margin is the average difference
between the standardized scores of matched positive and negative validation skills. Validation AUROC is computed from
the probabilities produced by a one-dimensional logistic classifier fitted to the standardized training scores. Figure~\ref{fig:pooling_analysis} presents the results for all six learned metrics.

\paragraph{Mean Paired Margin.}
As shown in Figure~\ref{fig:pooling_analysis}(a), the pooling strategy affects the separation between matched positive
and negative skills differently across metrics. Full-mean pooling produces larger margins for A2 and B1, increasing the
margin from 0.414 to 0.857 for A2 and from 0.397 to 1.459 for B1. It also gives a slightly larger margin for C2, with
0.816 compared with 0.768 under last-token pooling. In contrast, last-token pooling produces larger margins for B2, C1,
and D1. Their margins increase from 1.262 to 1.889, from 0.804 to 1.400, and from 0.328 to 1.162, respectively.
Averaged equally across the six metrics, the mean paired margin is 0.921 for full-mean pooling and 1.005 for last-token
pooling.

\paragraph{Validation AUROC.}
Figure~\ref{fig:pooling_analysis}(b) shows a similar metric-dependent pattern in ranking performance. Full-mean pooling
performs better for A2 and B1, achieving AUROCs of 0.731 and 0.926, compared with 0.621 and 0.610 under last-token
pooling. Last-token pooling achieves higher AUROC for the remaining four metrics. Specifically, it improves B2 from
0.973 to 1.000, C1 from 0.703 to 0.991, C2 from 0.722 to 0.749, and D1 from 0.632 to 0.881. Its macro-average AUROC is
0.809, compared with 0.781 for full-mean pooling.

These results indicate that pooling preference is metric dependent. Full-mean pooling is substantially stronger for A2
and B1, whereas last-token pooling provides better ranking performance for B2, C1, C2, and D1. C2 further shows that
the strategy with the larger paired margin does not necessarily achieve the higher AUROC, confirming that paired
separation and ranking performance capture complementary properties. Therefore, the results do not support a single
pooling strategy for all metrics and instead motivate selecting the pooling scope according to the semantic property
evaluated by each metric.

\section{Online and Cached-Vector Scoring Cost}
\label{sec:appendix_efficiency}

We measured the computational costs of online scoring and
scoring on cached vectors separately. The online experiment
used one NVIDIA H200 GPU and 60 unique validation documents,
with five positive and five negative documents per metric.
The workload comprised 24 short ($\leq 1{,}024$ tokens),
27 medium (1,025--4,096 tokens), and nine long
($>4{,}096$ tokens) documents. We fixed the batch size to one
and performed three technical repetitions per document
after warming up on five fixed documents. The model,
scoring parameters, and documents were preloaded.
Timing included tokenization, model inference, and scoring,
with GPU synchronization before and after each measurement.
Single-metric scoring required one encoder forward pass.
Scoring all six metrics required four forward passes,
as metrics with identical prefixes shared representation
extraction. We additionally measured CPU projection and
calibration on cached hidden vectors from the same documents
to isolate the cost of the scoring rule.

Table~\ref{tab:appendix_efficiency} summarizes the results.
SE-DEB achieved median online latencies of 212.24~ms for
a single metric and 852.09~ms for all six metrics, with
corresponding 95th-percentile latencies of 623.35~ms and
2,461.56~ms. The corresponding median latencies for SE-RAW
were 213.50~ms and 832.62~ms. Peak allocated GPU memory
was approximately 22.5--22.6~GiB for both methods.
With cached hidden vectors, median single-metric scoring
times were 2.50~$\mu$s for SE-RAW and 2.34~$\mu$s for SE-DEB.
These measurements exclude tokenization and model inference
and therefore quantify only the cost of scoring cached vectors.

These results indicate that representation extraction
dominates SkillEval's online cost, while projection and
calibration incur little additional overhead.
Sharing representation extraction reduces the number of
forward passes required for scoring all six metrics.
When hidden vectors are already available, the scoring
rule itself adds only a few microseconds of computation.

\begin{table*}[t]
\centering

\begin{tabular}{@{}llrrrrr@{}}
\toprule
\multicolumn{7}{@{}l}{
    \textbf{(a) Online scoring}
} \\
\midrule
Method & Workload & Median & p95 & Docs/s & Passes & Peak \\
       &          & (ms)   & (ms) &        &        & (GiB) \\
\midrule
SE-RAW & Single metric & 213.50 & 613.19
       & 3.56 & 1 & 22.54 \\
SE-RAW & Six metrics   & 832.62 & 2474.37
       & 0.89 & 4 & 22.61 \\
SE-DEB & Single metric & 212.24 & 623.35
       & 3.52 & 1 & 22.54 \\
SE-DEB & Six metrics   & 852.09 & 2461.56
       & 0.88 & 4 & 22.61 \\
\bottomrule
\end{tabular}

\medskip

\begin{tabular}{@{}lrr@{}}
\toprule
\multicolumn{3}{@{}l}{
    \textbf{(b) Single-metric scoring on cached vectors}
} \\
\midrule
Method & Median ($\mu$s) & p95 ($\mu$s) \\
\midrule
SE-RAW & 2.50 & 3.94 \\
SE-DEB & 2.34 & 3.00 \\
\bottomrule
\end{tabular}

\caption{
Online and cached-vector scoring costs of SkillEval.
Each workload uses 60 unique documents with three technical
repetitions per document.
(a) Online scoring uses one NVIDIA H200 GPU with batch
size one and includes tokenization, model inference,
and scoring. Model loading and document loading are excluded.
Docs/s is the number of document evaluations divided by
the total measured time. Passes denotes encoder forward
passes per document; Peak denotes peak allocated GPU memory.
(b) Cached-vector scoring measures CPU projection and
calibration only, excluding tokenization and model inference.
p95 denotes the 95th percentile of measured latency.
}
\label{tab:appendix_efficiency}
\end{table*}

\end{document}